\documentclass[conference]{IEEEtran}
\usepackage{amsmath,mathtools}
\usepackage{amssymb}
\usepackage{graphicx}
\usepackage{url}
\usepackage[numbers,square,sort&compress]{natbib}
\providecommand{\Description}[1]{}
\usepackage{booktabs,multirow,array,tabularx}
\usepackage{algorithm,algpseudocode}
\usepackage{microtype}
\usepackage{balance}
\usepackage{enumitem}
\usepackage{pifont}
\usepackage[numbers,square,sort&compress]{natbib}
\usepackage{pifont}
\usepackage{booktabs}
\usepackage{graphicx}

\usepackage[hidelinks]{hyperref}

\DeclareRobustCommand{\method}{\textsc{SkillCompiler}}

\newcommand{\CV}{\mathrm{CV}}

\newcommand{\para}[1]{\par\noindent\textbf{#1}\enspace}
\newcommand{\oppre}{\operatorname{pre}}
\newcommand{\oppost}{\operatorname{post}}
\newcommand{\opcheck}{\operatorname{check}}

\begin{document}
\title{Compile Away Uncertainty: Self-Evolving Agent Skills via Programmatic Compilation}
\author{
\IEEEauthorblockN{Chengyu Yang\textsuperscript{1}, Yunxiang Zheng\textsuperscript{1}, Zongzheng Li\textsuperscript{1}, Jiamei Li\textsuperscript{1}, Boxun Lan\textsuperscript{1}, Xiaoyu Liu\textsuperscript{2}, \\Xiaoyu Zhang\textsuperscript{1}, 
Xiaohong Chen\textsuperscript{1}, Ming Hu\textsuperscript{1,*}}
\IEEEauthorblockA{\textsuperscript{1}East China Normal University, \textsuperscript{2}China University of Geoscience}
\IEEEauthorblockA{\textsuperscript{*}Corresponding author}
}
\maketitle
\thispagestyle{plain} 
\pagestyle{plain}     
\begin{abstract}
Self-evolving agent skills enable AI agents to accumulate and refine reusable procedural knowledge from interaction experience without updating model parameters.
However, since such skills are typically represented as textual documents and executed through LLM reasoning, even previously learned procedures must be repeatedly interpreted and regenerated, introducing avoidable uncertainty into every execution. 
In contrast, program execution is deterministic and independently verifiable. 
Intuitively, transforming well-specified operations into executable programs can reduce reliance on LLM reasoning and, consequently, reduce execution uncertainty.
Motivated by this intuition, we introduce \method{}, a self-evolving skill framework that integrates programmatic compilation into the skill evolution process. 
Specifically, \method{} adopts an executability-aware skill compilation strategy that decomposes skills into fine-grained operations, analyzes their executability, and constructs hybrid skill skeletons that retain LLM reasoning for context-dependent operations while compiling well-specified ones into validated programs.
To avoid redundant program synthesis and enable executable knowledge to accumulate across evolving skills, we further design a program asset reuse mechanism that organizes validated programs as reusable assets, retrieves existing implementations whenever applicable, and synthesizes and verifies new programs only when necessary. 
Extensive experiments on five benchmarks and six model families demonstrate that SkillCompiler consistently outperforms state-of-the-art methods, achieving up to 32.33\% improvement in task performance.

\end{abstract}


\section{Introduction}\label{sec:introduction}

Large language model (LLM) agents have demonstrated strong capabilities in reasoning, planning, tool use, and environment interaction, enabling them to solve increasingly complex tasks, including information retrieval, Web interaction, and interactive problem solving~\citep{react,mind2web,webarena,webvoyager,autotools}. 
However, agents often need to rediscover similar task-solving procedures when facing new tasks, even when related experience has already been acquired. 
Agent skills provide a natural way to preserve such experience as reusable procedural knowledge, allowing agents to reuse previously acquired task-solving procedures without updating the underlying language model~\citep{reflexion,expel,fedskillgit,trace2skill,skillopt}. 
Consequently, effectively accumulating and reusing skills has become an important direction for improving agent capabilities through interaction experience.

To continuously accumulate and refine reusable skills from interaction experience, existing work has explored different strategies for skill evolution. 
Existing approaches can be broadly classified into two categories, i.e., skill content evolution and skill execution augmentation. Specifically, skill content evolution improves reusable procedural knowledge through trajectory distillation~\citep{trace2skill}, reflection-based revision~\citep{embodiskill}, and validation-guided skill optimization~\citep{skillopt}. 
Skill execution augmentation enhances how acquired skills are applied by reusing learned workflows~\citep{awm,flowevo}, incorporating external tool capabilities~\citep{autotools}, or constructing executable procedural representations~\citep{nsi,skilldisco,asi}. These methods enable agents to progressively improve both the knowledge encoded in skills and the way such knowledge is reused during task execution.


Although existing skill evolution methods can continuously accumulate and refine reusable task-solving knowledge, they still face two challenges in reliable and efficient skill reuse, i.e., execution uncertainty and repeated capability construction. 
Specifically, \ding{182} executing an evolved skill still requires the agent to repeatedly interpret retained knowledge, reconstruct intermediate decisions, and generate corresponding actions through LLM reasoning. As a result, even with an unchanged skill, repeated executions may follow different reasoning paths and produce different outcomes. 
\ding{183} Reusable execution capabilities discovered during skill evolution are often constructed for individual skills or tasks and are not persistently organized for subsequent reuse, leading to repeated generation and validation of similar procedures across evolving skills. 
Therefore, \textit{how to reduce execution uncertainty while efficiently accumulating and reusing executable knowledge across skill evolution remains an important problem.}

Typically, LLM reasoning is inherently probabilistic, which makes the execution of the same skill susceptible to different intermediate decisions and outcomes across repeated runs. 
In contrast, executable programs encode their inputs, control flow, and state transformations explicitly, so the same operation follows a fixed execution procedure under the same conditions rather than regenerating intermediate decisions at each invocation.
Intuitively, when an operation has explicit inputs, well-defined execution logic, and checkable outputs, its execution no longer needs to be repeatedly regenerated through LLM reasoning. 
Instead, such operations can be realized as executable programs, while LLM reasoning is retained for context-dependent or difficult-to-formalize decisions. 
Moreover, once these programs are validated, they can be persistently accumulated and reused across evolving skills, avoiding repeated synthesis of the same executable procedures.

Inspired by the above intuition, we propose \method{}, a self-evolving skill framework that transforms suitable reasoning operations into deterministic executable programs as skills evolve. 
Specifically, \ding{172} \method{} adopts an executability-aware skill compilation strategy to identify which skill operations can be reliably transformed into executable programs while retaining LLM reasoning for context-dependent decisions. \ding{173} To avoid redundant program synthesis and enable executable knowledge to accumulate across evolving skills, \method{} integrates a program asset reuse mechanism that organizes validated programs as reusable assets, retrieves compatible implementations whenever available, and synthesizes and verifies new programs only when necessary. In this way, \method{} reduces unnecessary LLM reasoning while preserving the flexibility required for unresolved operations.
The main contributions are summarized as follows.

\begin{itemize}
    \item We propose \method{}, a self-evolving skill framework that progressively transforms suitable skill operations into deterministic executable programs. Its executability-aware compilation strategy identifies which operations can be reliably programmatized while retaining LLM reasoning for context-dependent decisions.

    \item We design a program asset reuse mechanism that persistently accumulates, retrieves, and reuses validated executable programs across evolving skills, reducing redundant program synthesis and enabling executable knowledge to continuously grow through interaction.

    \item We conduct extensive experiments on five benchmarks and six model families. The results show that \method{} outperforms state-of-the-art methods, achieving up to 32.33\% improvement in task performance.
\end{itemize}


\section{Related Work}\label{sec:related}

\subsection{Experience Accumulation and Reusable Agent Skills}
AI agents can improve subsequent task solving by retaining useful experience from previous interactions without updating model parameters.
Reflexion~\citep{reflexion} stores verbal feedback from previous attempts to guide future decisions, while ExpeL~\citep{expel} extracts generalizable insights from collections of task experiences. 
Agent Workflow Memory~\citep{awm} induces recurring workflows from trajectories and retrieves them to guide subsequent action generation. 
Voyager~\citep{voyager} further accumulates executable skills from successful environment interactions and refines them through execution feedback.

\subsection{Agent Skills Evolution}

Existing work can be classified into \emph{skill content evolution} and \emph{skill execution augmentation}. 
Skill content evolution methods improve the quality and transferability of reusable procedural knowledge by refining skill representations from accumulated interaction experience. 
For example, Trace2Skill~\citep{trace2skill} distills reusable lessons from trajectories. 
EmbodiSkill~\citep{embodiskill} revises skills based on execution evidence.
SkillOpt~\citep{skillopt} conducts validation-guided optimization over natural-language skill documents. 
Skill execution augmentation methods enhance how acquired skills are carried out by introducing additional execution structures or capabilities. 
For example, Agent Workflow Memory~\citep{awm} reuses learned workflows to guide subsequent action generation. FlowEvo~\citep{flowevo} reuses verified workflows as executable skills. 
AutoTools~\citep{autotools} augments agent execution with external tools. 
NSI~\citep{nsi}, SkillDisCo~\citep{skilldisco}, and Agent Skill Induction~\citep{asi} construct more directly executable procedural representations from interaction experience.

Different from existing skill evolution methods, \method{} does not merely attach reusable workflows, tools, or executable artifacts to skill execution. Instead, it explicitly analyzes operation-level executability within an evolving skill and selectively determines which operations should remain under LLM reasoning and which should be transformed into validated programs. 
To the best of our knowledge, \method{} is the first self-evolving skill framework that jointly enables operation-level selective compilation and persistent reuse of the resulting program assets across skill evolution.

\begin{figure*}[t]
\centering
\includegraphics[width=0.9\textwidth]{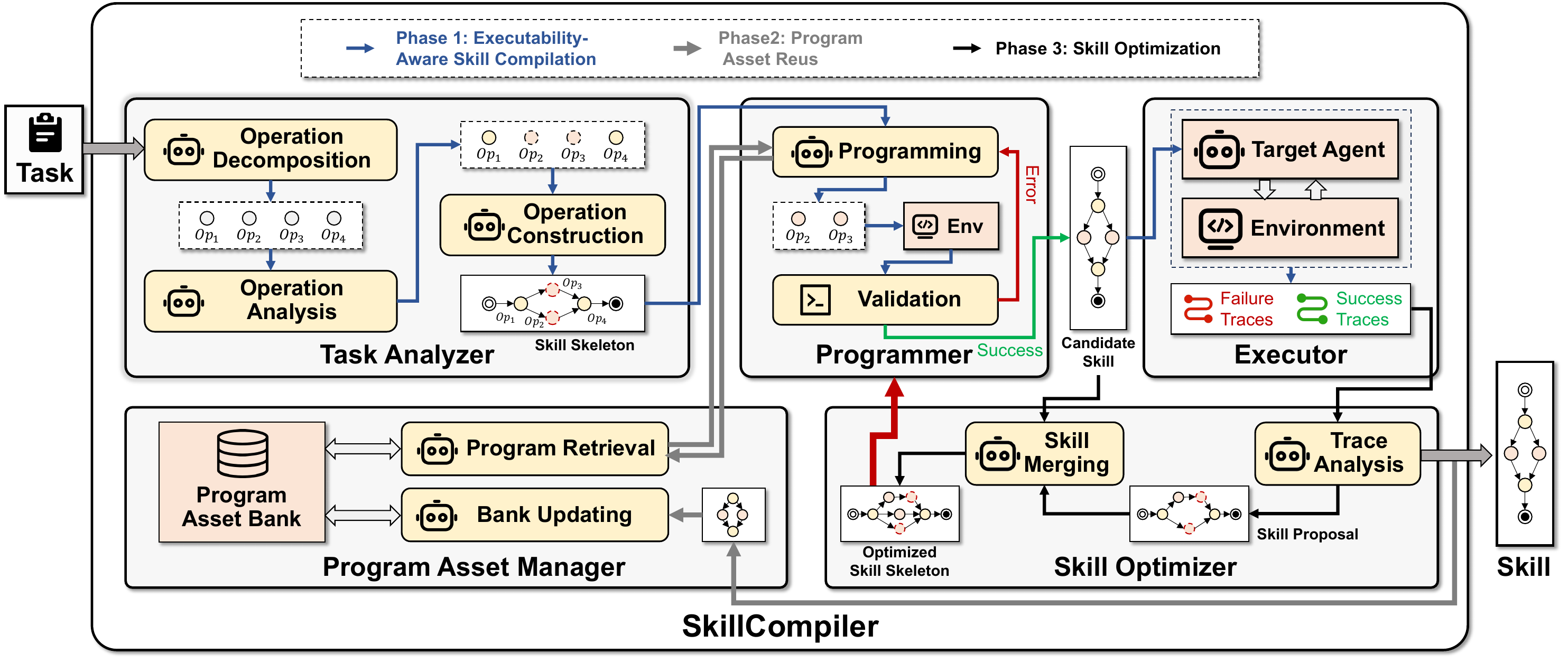}
\vspace{-0.1in}
\caption{Framework and Overview of \method}
\vspace{-0.2in}
\label{fig:overview}
\end{figure*}

\section{Our \method\ Approach}\label{sec:method}


\subsection{Overview}\label{sec:overview}

Fig.~\ref{fig:overview} presents the overall framework of \method{}, which consists of five major components, including the task analyzer, programmer, executor, skill optimizer, and program asset manager. 
Task analyzer decomposes a task into fine-grained operations, analyzes their executability, and constructs a hybrid skill skeleton. 
Programmer realizes programmable operations by retrieving existing implementations or synthesizing new programs and validates them before integration. 
Executor applies the resulting candidate skill by combining programmatic execution with decisions from the target agent. 
Skill optimizer analyzes successful and failed execution traces to refine the evolving skill. 
Program asset manager maintains a program asset bank to support program retrieval, updating, and reuse across skill evolution.

As shown in Fig.~\ref{fig:overview}, the workflow of \method{} mainly consists of three phases, i.e., executability-aware skill compilation, program asset reuse, and skill optimization. The details of each phase are as follows:
\begin{itemize}[leftmargin=*]
\item \textbf{Phase 1: Executability-aware skill compilation.} Given a task, the Task Analyzer decomposes it into fine-grained operations, analyzes their executability, and organizes both programmatic and agent-guided operations into a hybrid skill skeleton. For operations selected for programmatic execution, the Programmer obtains validated implementations and incorporates them into the skeleton to form a candidate skill, while the remaining operations continue to rely on LLM reasoning.

\item \textbf{Phase 2: Program asset reuse.} Before synthesizing a new implementation, the program asset manager retrieves compatible programs from the Program Asset Bank and reuses them whenever applicable. Newly synthesized and validated programs are persisted as reusable assets for subsequent skills.

\item \textbf{Phase 3: Skill optimization.} The executor applies the candidate skill in the target environment and records the resulting execution traces and task feedback. If the skill successfully completes the task, the execution result is returned and the validated skill knowledge is retained for subsequent reuse. Otherwise, the Skill Optimizer analyzes the failed execution trace to identify problematic operations, guidance, or dependencies and generates an updated skill skeleton. Operations affected by the revision are returned to the preceding phases for executability analysis, program retrieval or synthesis, and validation before the updated skill is executed again.
\end{itemize}

\subsection{Executability-Aware Skill Compilation}
\label{sec:compilation}

A skill may contain operations with different execution requirements. Some operations depend on semantic understanding, contextual judgment, or changing observations, while others follow explicit and repeatable procedures. Executing all operations through LLM reasoning therefore introduces unnecessary execution uncertainty, whereas compiling the entire skill into a program may sacrifice the flexibility required for context-dependent decisions. To address this issue, \method{} analyzes executability at the operation level and selectively transforms suitable operations into executable programs.

\subsubsection{Operation Decomposition.}
To identify the appropriate unit for selective compilation, the task analyzer first decomposes the task and relevant skill knowledge into a set of fine-grained operations. Each operation represents a locally coherent capability that can be independently invoked and evaluated. We represent an operation $\kappa_i $ using a 6-tuple operation contract as follows:
\begin{equation}
\kappa_i =
\langle I_i,O_i,\oppre_i,\oppost_i,\opcheck_i,g_i \rangle,
\end{equation}
where $I_i$ and $O_i$ denote the required inputs and expected outputs, $\oppre_i$ specifies the observable preconditions, $\oppost_i$ describes the expected local effects, $\opcheck_i$ defines how these effects can be verified, and $g_i$ contains reusable execution guidance. 
Task-specific entities and values are instantiated only when the operation is invoked, allowing the same operation to be reused across different tasks.

\subsubsection{Operation Analysis.}
After operation decomposition, the task analyzer determines whether each operation can be reliably realized through programmatic execution. An operation is considered programmatic when its inputs and outputs can be explicitly represented, its execution procedure can be clearly specified, its preconditions can be observed from the current environment, and its resulting outputs or local effects can be independently verified. In addition, the available tool interfaces should support a bounded implementation of the operation. Operations satisfying these requirements are selected for programmatic execution, while operations involving unresolved inputs, context-dependent decisions, or open-ended semantic reasoning remain agent-guided.

This analysis determines the execution mode of each operation before constructing the skill. In particular, programmatic operations no longer require the target agent to repeatedly generate their internal execution procedures, while agent-guided operations retain LLM reasoning to accommodate dynamic observations and contextual decisions. In this way, \method{} selectively moves well-specified operations from probabilistic reasoning to deterministic execution without restricting the flexibility of the entire skill.

\subsubsection{Skill Skeleton Construction.}
Based on the executability analysis, the task analyzer organizes the resulting operations into a hybrid skill skeleton. For a task $x$, the instantiated skill skeleton is represented as
\begin{equation}
S_x =
\langle\mathcal O_x,\sigma_x,\mathcal E_x,\beta_x\rangle,
\end{equation}
where $\mathcal O_x$ contains the operation occurrences required by the task, $\sigma_x$ specifies their execution order, $\mathcal E_x$ records the data and prerequisite relations among operations, and $\beta_x$ contains the current argument bindings. Data relations connect the outputs of preceding operations to the inputs of subsequent operations, while prerequisite relations specify the local effects that must be established before an operation can be executed.

Both programmatic and agent-guided operations are organized in the same skill skeleton. During construction, \method checks the compatibility between operation interfaces and identifies the sources of required inputs.
Task-specific values that are unavailable during construction remain unresolved in $\beta_x$ and are instantiated according to actual observations during execution. Therefore, the skill skeleton explicitly describes how deterministic programs and LLM reasoning collaborate to accomplish a task while preserving their respective execution requirements.

\subsubsection{Programming and Validation.}
For each operation selected for programmatic execution, \method needs to obtain a qualified executable realization before incorporating it into the candidate skill. The programmer first attempts to obtain an existing compatible implementation through the program asset manager. When no qualified implementation is available, it synthesizes a new program according to the operation contract, permitted tool interfaces, and available execution evidence.

The synthesized program is validated in an isolated environment before being used for task execution. Specifically, \method checks whether the program can be successfully executed, whether its inputs and outputs conform to $I_i$ and $O_i$, and whether the resulting local effects satisfy $\oppost_i$ according to $\opcheck_i$. When validation fails, the observed errors and check discrepancies are returned to the Programmer to guide program repair. This synthesis, validation, and repair process continues until a qualified implementation is obtained or the operation remains agent-guided. Only validated programs are incorporated into the skill skeleton to form the candidate skill, ensuring that LLM reasoning is replaced only when the corresponding program can be independently verified.

\vspace{-1mm}
\subsection{Program Asset Reuse}
\label{sec:reuse}

Although programmatic execution reduces repeated LLM reasoning, repeatedly synthesizing and validating similar programs can introduce additional overhead during skill evolution. In practice, different tasks may contain operations with the same underlying capability, making their executable implementations reusable across skills. To avoid redundant program synthesis, \method{} maintains validated programs in the Program Asset Bank and preferentially reuses existing implementations whenever compatible operations are encountered.

\subsubsection{Compatibility-Based Program Retrieval.}
For each operation selected for programmatic execution, the program asset manager first retrieves candidate programs according to the semantic description and execution requirements of the operation. 
Lexical similarity is used to identify potentially reusable implementations, while their applicability is further determined according to the operation contract. 
Specifically, compatible Skill Skeletons are ranked by lexical similarity, while atomic skills are ranked using lexical Dice similarity, historical utility, and activation status:
\begin{equation}
\label{eq:retrieval_score}
R(a,o) =
\frac{2|\mathcal{T}(o)\cap\mathcal{T}(a)|}
{|\mathcal{T}(o)|+|\mathcal{T}(a)|}
+0.1\,u(a)
+0.05\,\mathbf{1}[\mathrm{active}(a)],
\end{equation}
where $\mathcal{T}(\cdot)$ denotes lexical term sets,
$u(a)$ is the historical utility of atomic skill $a$,
and $\mathbf{1}[\mathrm{active}(a)]$ indicates its active status. Candidates are first filtered by capability and contract compatibility before ranking.

Program realizations are located through explicit Skill-Implementation-Program references.
A retrieved program can be reused only when its required inputs can be correctly bound, its preconditions are satisfied, the current environment provides the required interfaces, and its existing validation record remains applicable to the current operation. Formally, the eligible programs are
\begin{equation}
\label{eq:program_compatibility}
\begin{aligned}
\mathcal{P}_{\mathrm{valid}}(a,x)
=\{p\in\mathcal{P}(a):\;&
\mathrm{Bind}(p,x)\land\mathrm{Pre}(p,x)\\
&\land\mathrm{Env}(p,x)\land\mathrm{Cert}(p,x)\},
\end{aligned}
\end{equation}
where $\mathcal{P}(a)$ contains the program realizations referenced by skill $a$, and the predicates check argument binding, preconditions, environment compatibility, and validation applicability, respectively.
Therefore, lexical similarity supports candidate discovery, while interface and state compatibility determine whether a retrieved program can be safely invoked.

When multiple implementations correspond to the same capability, \method{} retains their respective applicability conditions and validation evidence. 
The program satisfying the current operation contract and execution environment is selected for reuse. Automatic invocation requires a unique eligible or preferred implementation; otherwise, selection remains agent-guided. 
If no compatible implementation is available, the operation is passed to the Programmer for program synthesis and validation as described in Section~\ref{sec:compilation}.

\subsection{Skill Optimization}
\label{sec:updating}

After constructing the candidate skill, \method{} executes it in the target environment to evaluate whether the current skill can successfully accomplish the task. During execution, the executor invokes validated programs for programmatic operations and delegates the remaining operations to the target agent. The resulting actions, observations, argument bindings, program invocations, local checking results, and final task feedback are recorded as execution traces for subsequent skill updating.

\subsubsection{Trace-Based Skill Revision.}
When task execution fails, the skill optimizer locates the failure by comparing the actual execution of each operation with its operation contract. If an operation produces the expected output and satisfies $\opcheck_i$, the operation is retained even when the final task fails. Otherwise, \method{} revises the corresponding component according to the observed failure. An unsatisfied or overly broad invocation condition leads to an update of $\oppre_i$. Incorrect information passed between operations leads to changes in argument bindings or dependencies in the skill skeleton. Incomplete execution guidance results in an update of $g_i$. If a program fails to produce the expected effect specified by $\oppost_i$, the corresponding program is sent back for repair and validation. Therefore, skill revision is performed locally according to operation-level execution evidence rather than regenerating the entire skill from the failed task.

\subsubsection{Skill Merging and Updating.}
The revised operations are further compared with existing operations in the skill bank to reduce redundant skill knowledge. \method{} first determines whether two operations describe the same local capability and then checks whether their operation contracts are compatible. Operations with consistent input and output semantics, preconditions, expected effects, and checking requirements are merged into a shared reusable operation, while task-specific entities and values are abstracted through parameterization and argument bindings. Operations with different capability semantics or conflicting contracts remain separate.

Formally, for two operation contracts $\kappa_i$ and $\kappa_j$, the merging criterion is expressed as
\begin{equation}
\label{eq:skill_merge}
\begin{aligned}
&\mathrm{Mergeable}(\kappa_i,\kappa_j)
= \mathrm{SameCap}(\kappa_i,\kappa_j)\\
&\quad \land\ \exists\,\pi:
\bigwedge_{f\in\{I,O,\mathrm{pre},\mathrm{post},\mathrm{check}\}}
\mathrm{Compat}\bigl(\pi(f_i),f_j\bigr).
\end{aligned}
\end{equation}
Here, $\mathrm{SameCap}$ denotes shared capability semantics, $\pi$ aligns task-specific parameters, and $\mathrm{Compat}$ requires semantically consistent contract fields under the alignment. Thus, merging is permitted only when capability identity and contract compatibility are both satisfied.

The resulting operations are reorganized into an updated skill skeleton. If the update only changes agent-guided operations, the revised skeleton can be directly used for the next execution. 
If the update changes a programmatic operation, its implementation is retrieved again or regenerated when necessary and subsequently validated before reuse. 
The candidate skill is then executed in the target environment. A successful execution confirms the current skill update with corresponding skill and program resources retained, while another failure produces a new execution trace for the next round of local revision.

\begin{table}[!b]
\centering
\caption{Dataset partition statistics for the five evaluation benchmarks.}
\label{tab:splits}
\small
\renewcommand{\arraystretch}{1.00}
\begin{tabular*}{\linewidth}{@{}l@{\extracolsep{\fill}}rrr@{}}
\toprule
\textbf{Benchmark} & \textbf{Train} & \textbf{Val} & \textbf{Test}\\
\midrule
SearchQA~\citep{searchqa} & 300 & 24 & 1,400\\
OfficeQA~\citep{officeqa} & 120 & 24 & 102\\
DocVQA~\citep{docvqa}   & 180 & 22 & 201\\
LiveMath~\citep{livemath} & 60  & 17 & 100\\
ALFWorld~\citep{alfworld} & 120 & 24 & 134\\
\bottomrule
\end{tabular*}
\vspace{-5mm}
\end{table}

\begin{table*}[!t]
\centering
\caption{Performance comparison of \method\ and baseline methods across four benchmarks using six models.}
\label{tab:main}
\footnotesize
\renewcommand{\arraystretch}{1.00}
\begin{tabular*}{\textwidth}{l|l|@{\extracolsep{\fill}}cccc|c@{}}
\toprule
\textbf{Model} & \textbf{Method} & \textbf{SearchQA} & \textbf{OfficeQA} & \textbf{DocVQA} & \textbf{LiveMath} & \textbf{Avg}\\
\midrule
\multirow{8}{*}{MiniMax-M3}
 & No Skill
 & 76.21 & 52.94 & 72.64 & 32.00 & 58.45\\
 & Human Skill
 & 76.17 & 53.92 & 83.91 & 33.67 & 61.92\\
 & Trace2Skill~\citep{trace2skill}
 & 79.43 & 60.78 & 91.38 & 42.00 & 68.40\\
 & EvoSkill~\citep{evoskill}
 & 79.24 & 62.09 & 89.88 & 51.33 & 70.64\\
 & EmbodiSkill~\citep{embodiskill}
 & 77.64 & 61.76 & 79.77 & 34.67 & 63.46\\
 & GEPA~\citep{gepa}
 & 79.57 & 59.48 & 90.55 & 48.67 & 69.57\\
 & SkillOpt~\citep{skillopt}
 & 81.88 & 60.46 & 89.72 & 54.67 & 71.68\\
 & \textbf{\method}
 & \textbf{82.86} & \textbf{63.72} & \textbf{94.53} & \textbf{78.00} & \textbf{79.78}\\
\midrule
\multirow{8}{*}{Gemini-3.8-Flash}
 & No Skill
 & 82.83 & 50.98 & 92.70 & 33.00 & 64.88\\
 & Human Skill
 & 83.19 & 51.63 & 94.86 & 33.33 & 65.75\\
 & Trace2Skill~\citep{trace2skill}
 & 84.67 & 60.78 & 94.20 & 45.33 & 71.25\\
 & EvoSkill~\citep{evoskill}
 & 84.74 & 57.84 & 94.36 & 55.67 & 73.15\\
 & EmbodiSkill~\citep{embodiskill}
 & 82.55 & 25.49 & 90.71 & 33.00 & 57.94\\
 & GEPA~\citep{gepa}
 & 87.05 & 66.34 & 93.70 & 67.33 & 78.61\\
 & SkillOpt~\citep{skillopt}
 & 84.29 & 55.23 & 94.69 & 66.67 & 75.22\\
 & \textbf{\method}
 & \textbf{89.00} & \textbf{75.49} & \textbf{96.02} & \textbf{78.00} & \textbf{84.63}\\
\midrule
\multirow{8}{*}{Qwen3.8-Flash}
 & No Skill
 & 79.10 & 37.58 & 93.53 & 25.67 & 58.97\\
 & Human Skill
 & 79.14 & 50.98 & 93.86 & 26.00 & 62.50\\
 & Trace2Skill~\citep{trace2skill}
 & 82.90 & 72.88 & 93.70 & 31.67 & 70.29\\
 & EvoSkill~\citep{evoskill}
 & 82.74 & 70.59 & 93.53 & 51.67 & 74.63\\
 & EmbodiSkill~\citep{embodiskill}
 & 83.07 & 67.65 & 94.69 & 14.67 & 65.02\\
 & GEPA~\citep{gepa}
 & 81.21 & 68.63 & 93.37 & 39.00 & 70.55\\
 & SkillOpt~\citep{skillopt}
 & 80.21 & 68.30 & 92.70 & 35.00 & 69.05\\
 & \textbf{\method}
 & \textbf{85.07} & \textbf{75.49} & \textbf{96.02} & \textbf{77.00} & \textbf{83.40}\\
\midrule
\multirow{8}{*}{DeepSeek-V4.1-Flash}
 & No Skill
 & 76.10 & 33.33 & 74.96 & 28.00 & 53.10\\
 & Human Skill
 & 76.21 & 41.18 & 82.75 & 27.33 & 56.86\\
 & Trace2Skill~\citep{trace2skill}
 & 84.24 & 67.65 & 88.56 & 32.67 & 68.28\\
 & EvoSkill~\citep{evoskill}
 & 83.33 & 69.28 & 84.58 & 52.00 & 72.30\\
 & EmbodiSkill~\citep{embodiskill}
 & 81.36 & 51.63 & 77.61 & 23.67 & 58.57\\
 & GEPA~\citep{gepa}
 & 83.21 & 70.26 & 87.56 & 50.67 & 72.93\\
 & SkillOpt~\citep{skillopt}
 & 82.95 & 69.93 & 85.57 & 30.33 & 67.20\\
 & \textbf{\method}
 & \textbf{86.36} & \textbf{71.24} & \textbf{91.71} & \textbf{77.00} & \textbf{81.58}\\
\midrule
\multirow{8}{*}{Kimi-K2.6}
 & No Skill
 & 80.79 & 32.03 & 94.67 & 39.67 & 61.79\\
 & Human Skill
 & 81.05 & 32.03 & 95.02 & 39.33 & 61.86\\
 & Trace2Skill~\citep{trace2skill}
 & 83.98 & 55.56 & 95.36 & 43.33 & 69.56\\
 & EvoSkill~\citep{evoskill}
 & 83.33 & 52.61 & 95.19 & 36.00 & 66.78\\
 & EmbodiSkill~\citep{embodiskill}
 & 78.98 & 57.52 & 92.37 & 51.67 & 70.14\\
 & GEPA~\citep{gepa}
 & 83.10 & 55.56 & 96.02 & 42.00 & 69.17\\
 & SkillOpt~\citep{skillopt}
 & 81.48 & 54.25 & 95.19 & 54.00 & 71.23\\
 & \textbf{\method}
 & \textbf{85.50} & \textbf{74.51} & \textbf{96.52} & \textbf{76.00} & \textbf{83.13}\\
\midrule
\multirow{8}{*}{GLM-5.3-Flash}
 & No Skill
 & 78.62 & 30.07 & 81.92 & 42.00 & 58.15\\
 & Human Skill
 & 79.45 & 31.70 & 91.04 & 41.00 & 60.80\\
 & Trace2Skill~\citep{trace2skill}
 & 83.90 & 65.69 & 94.36 & 44.67 & 72.16\\
 & EvoSkill~\citep{evoskill}
 & 77.38 & 59.80 & 90.38 & 40.67 & 67.06\\
 & EmbodiSkill~\citep{embodiskill}
 & 81.02 & 59.15 & 93.20 & 40.00 & 68.34\\
 & GEPA~\citep{gepa}
 & 83.50 & 64.05 & 93.03 & 43.00 & 70.90\\
 & SkillOpt~\citep{skillopt}
 & 83.55 & 64.38 & 93.37 & 41.67 & 70.74\\
 & \textbf{\method}
 & \textbf{85.21} & \textbf{71.89} & \textbf{94.53} & \textbf{77.00} & \textbf{82.16}\\
\bottomrule
\end{tabular*}
\end{table*}

\section{Experiments}\label{sec:experiments}
In this section, we first introduce the experimental setup and evaluate the performance of \method\ against representative baselines. We then investigate its execution reliability and resource consumption through repeated-execution experiments and cost analysis. Finally, ablation studies are conducted to assess the contributions of the key components.

\subsection{Experimental Setup}
\label{sec:setup}

\noindent\textbf{Benchmarks and Models.}
We conduct experiments on five benchmarks, including SearchQA~\citep{searchqa}, OfficeQA~\citep{officeqa}, DocVQA~\citep{docvqa}, LiveMath~\citep{livemath}, and ALFWorld~\citep{alfworld}.
Specifically, SearchQA~\citep{searchqa} evaluates open-domain question answering, while OfficeQA~\citep{officeqa} and DocVQA~\citep{docvqa} focus on document understanding.
LiveMath~\citep{livemath} assesses mathematical reasoning, and ALFWorld~\citep{alfworld} evaluates long-horizon interactive task solving.
Table~\ref{tab:splits} summarizes the training, validation, and test partitions of these benchmarks.
Furthermore, we consider six large language models, including MiniMax-M3, Gemini-3.8-Flash, Qwen3.8-Flash, DeepSeek-V4.1-Flash, Kimi-K2.6, and GLM-5.3-Flash.

\noindent\textbf{Baselines.}
We compare our proposed method \method\ with seven baselines, including No Skill, Human Skill, Trace2Skill~\citep{trace2skill}, EvoSkill~\citep{evoskill}, EmbodiSkill~\citep{embodiskill}, GEPA~\citep{gepa}, and SkillOpt~\citep{skillopt}.
\begin{itemize}
    \item \textbf{No Skill.} The agent performs tasks without any skill guidance, relying solely on the underlying LLM.
    \item \textbf{Human Skill.} The agent executes tasks using fixed human-written skills provided by SkillOpt without evolution.
    \item \textbf{Trace2Skill}~\citep{trace2skill}.  It distills reusable knowledge from execution trajectories and consolidates trajectory-level lessons into transferable skills.
    \item \textbf{EvoSkill}~\citep{evoskill}. It identifies capability gaps through failure analysis and iteratively generates or refines skills based on validation feedback.
    \item \textbf{EmbodiSkill}~\citep{embodiskill}. It employs skill-aware reflection to distinguish skill deficiencies from execution failures, enabling targeted skill revisions.
    \item \textbf{GEPA}~\citep{gepa}. It optimizes prompts through natural-language reflection on execution feedback and Pareto-based candidate selection.
    \item \textbf{SKillOpt}~\citep{skillopt}. It iteratively optimizes natural-language skills through bounded textual revisions, accepting updates only when validation performance improves.
\end{itemize}

\noindent\textbf{Evaluation Protocol.}
All methods use the same task partitions, available tools, benchmark evaluators, and inference settings within each model--benchmark configuration. Main performance and cost results are averaged over three independently seeds (42, 43, and 44). Persistent skill states are selected before testing and remain frozen throughout the test evaluation. For repeated-execution and ablation analyses, we use the seed-42 skill states and execute each test task three times under the same inputs and environment conditions.

\noindent\textbf{Metrics.}
Task performance is evaluated using exact match (EM) where applicable and benchmark-specific scores otherwise, including success rate (SR) for ALFWorld.
Execution reliability is assessed through outcome instability and execution-path consistency, while resource consumption is measured by model token usage and the number of model calls.

\subsection{Performance Comparison}
\label{sec:main}
We evaluate the effectiveness of \method\ across different benchmarks, large language model families, and long-horizon interaction tasks.

\subsubsection{Performance on different benchmarks}

Table~\ref{tab:main} presents the performance comparison across six language model families in terms of EM on four benchmarks.
Taking Kimi-K2.6 as an example, \method\ achieves scores of 85.50\%, 74.51\%, 96.52\%, and 76.00\% on SearchQA, OfficeQA, DocVQA, and LiveMath, respectively.
Compared with the strongest baseline on each benchmark, these results correspond to improvements of 1.52\%, 16.99\%, 0.50\%, and 22.00\%, respectively.
In particular, the most significant improvements are observed on LiveMath and OfficeQA, indicating that \method\ is effective on mathematical reasoning and document-oriented tasks.
In contrast, the relatively modest gains on SearchQA and DocVQA suggest that the benefits of programmatic compilation vary across task characteristics.
A similar trend is observed across the remaining model families, with the largest performance gains generally occurring on LiveMath.
In particular, \method\ improves the performance of GLM-5.3-Flash on LiveMath by 32.33\%, further highlighting its effectiveness on mathematical reasoning tasks.
These results suggest that \method\ is advantageous for tasks involving well-specified operations that can be compiled into validated executable programs.

\subsubsection{Performance with different model families}
To investigate the applicability of \method\ across different foundation models, we evaluate its performance using six language model families.
As shown in Table~\ref{tab:main}, \method\ consistently achieves the highest average score across all six models, outperforming the strongest baseline for each model by 6.02--11.90\%.
Specifically, the average scores of \method\ range from 79.78\% to 84.63\%, exhibiting a relatively narrow performance range of 4.85\% across different models.
The improvement is observed even though the relative performance of competing skill evolution methods changes across models. For example, the strongest overall baseline differs between Gemini-3.8-Flash and Qwen3.8-Flash, whereas \method\ remains the top-performing method in both settings. This result suggests that the proposed skill representation and execution mechanism can be applied across heterogeneous model families without modifying their parameters. It does not imply identical performance across models; rather, the advantages over the corresponding baselines are consistently positive.
These results indicate that \method\ maintains consistent performance advantages across different foundation models, demonstrating its broad applicability without modifying the underlying model parameters.

\begin{table}[!t]
    \centering
    \caption{Performance comparison on ALFWorld using DeepSeek-V4.1-Flash.}
    \label{tab:alfworld_results}
    \begin{tabularx}{\linewidth}{
        >{\raggedright\arraybackslash}X
        >{\centering\arraybackslash}X
    }
        \toprule
        \textbf{Method} & \textbf{Success Rate (\%)} \\
        \midrule
        No Skill       & 79.60 \\
        Human Skill    & 83.83 \\
        Trace2Skill    & 88.31 \\
        EvoSkill       & 85.82 \\
        EmbodiSkill    & 93.53 \\
        GEPA           & 86.82 \\
        SkillOpt       & 88.06 \\
        \textbf{\method} & \textbf{98.26} \\
        \bottomrule
    \end{tabularx}
    \vspace{-5mm}
\end{table}

\subsubsection{Performance on long-horizon interaction tasks}
To further evaluate the effectiveness of \method\ in long-horizon interactive environments, we conduct experiments on ALFWorld using DeepSeek-V4.1-Flash.
Unlike the other four benchmarks, ALFWorld requires agents to perform sequential actions and adapt to changing environment states, where intermediate decisions and execution errors may affect subsequent task progress.
As shown in Table~\ref{tab:alfworld_results}, \method\ achieves the highest success rate of 98.26\%, outperforming all competing methods.
Specifically, \method\ surpasses the strongest baseline EmbodiSkill by 4.73\%, while exceeding SkillOpt and GEPA by 10.20\% and 11.44\%, respectively.
Compared with No Skill, \method\ improves the success rate from 79.60\% to 98.26\%, corresponding to an improvement of 18.66\%.
These results demonstrate the effectiveness of \method\ in long-horizon interactive tasks, indicating that its performance advantages extend beyond conventional reasoning and question-answering tasks to complex long-horizon interaction task.

\begin{figure*}[!t]
\centering
\includegraphics[width=\textwidth]{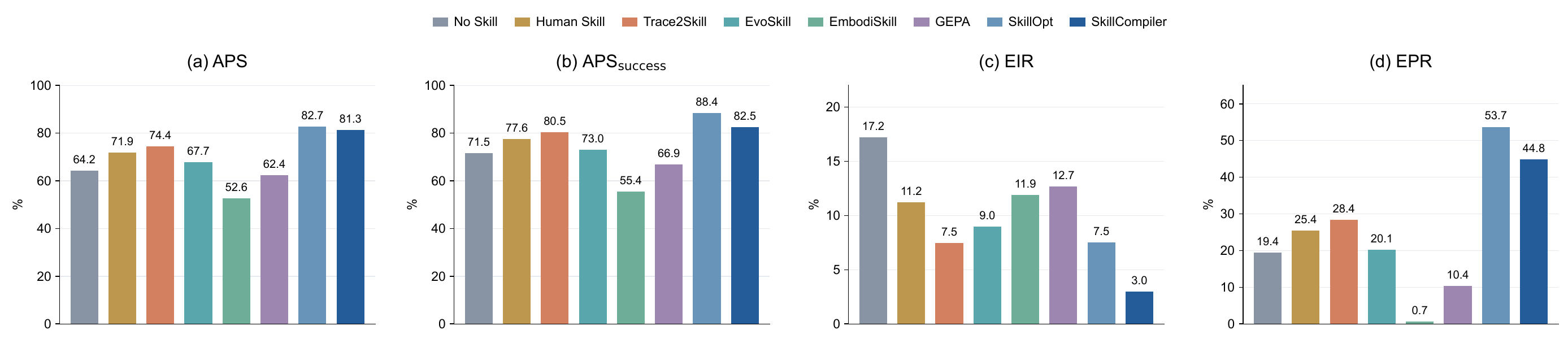}
\vspace{-0.3in}
\caption{Execution reliability comparison on ALFWorld using DeepSeek-V4.1-Flash.}
\label{fig:alfworld_reliability}
\Description{Four panels compare No Skill, Human Skill,
SkillOpt, EmbodiSkill, GEPA, and SkillCompiler on average
pairwise path similarity, success-conditioned path similarity,
outcome flips, and exact path agreement. \method\ has
the lowest EIR; SkillOpt has the highest path agreement.}
\label{fig:stability}
\vspace{-0.1in}
\end{figure*}

\subsection{Uncertainty Analysis}
\begin{table}[!t]
\centering
\caption{Performance and execution uncertainty comparison on LiveMath using Gemini-3.8-Flash.}
\label{tab:livemath_stability}
\small
\setlength{\tabcolsep}{3pt}
\renewcommand{\arraystretch}{1.00}
\begin{tabular*}{\linewidth}{@{}l@{\extracolsep{\fill}}rr@{}}
\toprule
\textbf{Method} & \textbf{EM $\uparrow$} & \textbf{EIR@3 $\downarrow$}\\
\midrule
No Skill    & 28.33 & 19.00\\
Human Skill & 28.33 & 13.00\\
Trace2Skill & 40.67 & 28.00\\
EvoSkill    & 55.33 & 31.00\\
EmbodiSkill & 27.67 & 18.00\\
GEPA        & 58.00 & 23.00\\
SkillOpt    & 64.67 & 32.00\\
\textbf{\method}     & \textbf{78.67} & \textbf{11.00}\\
\bottomrule
\end{tabular*}
\vspace{-5mm}
\end{table}
To investigate the execution uncertainty, we execute each task three times using fixed learned skills under identical task inputs, initial environment states, and inference settings.
We adopt execution instability rate (EIR@3) to measure the proportion of tasks exhibiting both successful and failed executions.
For ALFWorld, we further employ average path similarity (APS@3), success-conditioned path similarity (APS$_{\mathrm{succ}}$@3), and exact path repetition (EPR@3) to evaluate execution-path consistency.
Table~\ref{tab:livemath_stability} presents the results on LiveMath using Gemini-3.8-Flash.
\method\ achieves the highest EM of 78.67\% and the lowest EIR of 11.00\%.
Compared with SkillOpt, which achieves the highest EM among the baselines, \method\ improves EM by 14.00\% while reducing EIR from 32.00\% to 11.00\%, corresponding to a relative reduction of 65.6\%.
Similarly, \method\ achieves the lowest EIR of 3.0\% on ALFWorld, outperforming SkillOpt by 4.5\%, as shown in Fig.~\ref{fig:alfworld_reliability}.
Regarding execution-path consistency, \method\ achieves APS, APS$_{\mathrm{succ}}$, and EPR scores of 81.3\%, 82.5\%, and 44.8\%, respectively.
These results demonstrate that \method\ effectively reduces execution uncertainty across different task environments while maintaining high task performance.

\subsection{Cost Analysis}
\label{sec:cost}
To evaluate the resource efficiency of \method, we analyze the learning and inference costs in terms of token consumption and model calls.
In particular, learning costs include tokens consumed during skill evolution, while inference costs are measured by the average token consumption and model calls per test task.
The results are reported in Tables~\ref{tab:cost_livemath} and~\ref{tab:cost_alfworld}.
Specifically, on LiveMath, \method\ consumes 2.695M learning tokens, reducing the learning cost by 41.9\% and 52.1\% compared with SkillOpt and GEPA, respectively.
Moreover, \method\ requires only 0.450 model calls per task, the lowest among all competing methods.
Although its inference token consumption of 13.984k per task exceeds that of SkillOpt, it remains lower than those of EvoSkill, EmbodiSkill, and GEPA.
On ALFWorld, \method\ consumes 47.015k inference tokens per task, representing reductions of 7.6\% and 32.7\% compared with SkillOpt and GEPA, respectively.
Meanwhile, its learning cost of 32.442M tokens is also lower than those of both methods.
Compared with No Skill, \method\ reduces model calls by 25.7\% while improving the task success rate by 18.66\%.
These results indicate that \method\ achieves substantial performance improvements without necessarily increasing inference costs.
Furthermore, Table~\ref{tab:program_reuse} reports the internal execution and program reuse statistics of \method\ on ALFWorld.
The PPC reaches 100.00\%, while CPO is at least 48.36\%.
Meanwhile, CSR, PIS, and PAS achieve 54.85\%, 88.84\%, and 87.06\%, respectively.
In particular, the high PAS indicates that a substantial proportion of environment actions are performed within program execution spans.
These observations provide additional evidence of extensive programmatic execution, complementing the resource efficiency analysis.

\begin{table}[!t]
\centering
\caption{Resource consumption comparison on LiveMath using Gemini-3.8-Flash.}
\label{tab:cost_livemath}
\footnotesize
\setlength{\tabcolsep}{3pt}
\renewcommand{\arraystretch}{1.00}
\begin{tabular*}{\linewidth}{@{}l@{\extracolsep{\fill}}rrr@{}}
\toprule
\textbf{Method} 
       & \textbf{Learning (M tokens)}
       & \textbf{Test/task (k tokens)}
       & \textbf{Calls/task}\\
\midrule
No Skill    & 0.000 & 5.429  & 1.007\\
Human Skill & 0.000 & 5.997  & 1.000\\
Trace2Skill & 1.117 & 8.918  & 1.003\\
EvoSkill    & 1.752 & 17.311 & 1.060\\
EmbodiSkill & 2.988 & 19.335 & 3.047\\
GEPA        & 5.631 & 17.504 & 1.033\\
SkillOpt    & 4.639 & 6.672  & 1.007\\
\textbf{\method}     & 2.695 & 13.984 & \textbf{0.450}\\
\bottomrule
\end{tabular*}
\end{table}

\begin{table}[!t]
\centering
\vspace{-0.1in}
\caption{Resource consumption comparison on ALFWorld using DeepSeek-V4.1-Flash.}
\vspace{-0.1in}
\label{tab:cost_alfworld}
\footnotesize
\setlength{\tabcolsep}{3pt}
\renewcommand{\arraystretch}{1.00}
\begin{tabular*}{\linewidth}{@{}l@{\extracolsep{\fill}}rrr@{}}
\toprule
\textbf{Method} 
       & \textbf{Learning (M tokens)}
       & \textbf{Test/task (k tokens)}
       & \textbf{Calls/task}\\
\midrule
No Skill    & 0.000  & 47.906 & 20.020\\
Human Skill & 0.000  & 27.885 & 16.813\\
Trace2Skill & 3.212  & 68.418 & 14.085\\
EvoSkill    & 3.515  & 39.102 & 19.420\\
SkillOpt    & 57.468 & 50.906 & 14.674\\
EmbodiSkill & 10.934 & 42.710 & 15.321\\
GEPA        & 52.970 & 69.836 & 16.279\\
\textbf{\method}     & 32.442 & 47.015 & 14.868\\
\bottomrule
\end{tabular*}
\vspace{-5mm}
\end{table}

\begin{table}[t]
    \centering
    \vspace{-0.1in}
    \caption{Program execution and reuse statistics of \method\ on ALFWorld across three learning seeds.}
    \label{tab:program_reuse}
    \vspace{-0.1in}
    \resizebox{\linewidth}{!}{
        \begin{tabular}{@{}llr@{}}
            \toprule
            \textbf{Metric} & \textbf{Definition} & \textbf{Mean (\%)} \\
            \midrule
            PPC & Persistent Program Coverage (covered / total tasks) & 100.00 \\
            CPO & Consumed Program Outputs (trace-confirmed lower bound) & $\geq 48.36$ \\
            CSR & Cross-Skill Reuse (multi-skill / used programs) & 54.85 \\
            PIS & Program Invocation Success (successful / total calls) & 88.84 \\
            PAS & Program Action Share (program / total actions) & 87.06 \\
            \bottomrule
        \end{tabular}
    }
\end{table}

\subsection{Ablation Study}\label{sec:ablation}

\begin{table}[!htbp]
\centering
\vspace{-0.1in}
\caption{Ablation study on ALFWorld using DeepSeek-V4.1-Flash.}
\vspace{-0.1in}
\label{tab:ablation}
\footnotesize
\setlength{\tabcolsep}{3pt}
\renewcommand{\arraystretch}{1.00}
\begin{tabular*}{\linewidth}{@{}l@{\extracolsep{\fill}}rrrr@{}}
\toprule
\textbf{Condition} &\textbf{ SR $\uparrow$}
          & \textbf{\shortstack[r]{Test tokens\\(k/task) $\downarrow$}}
          & \textbf{EIR@3 $\downarrow$}
          & \textbf{APS@3 $\uparrow$}\\
\midrule
Full \method
    & \textbf{98.26} & \textbf{41.954}
    & \textbf{2.99} & \textbf{81.27}\\
w/o Skill Skeleton
    & 95.52 & 115.424 & 8.21 & 67.49\\
w/o Programmer
    & 53.48 & 308.887 & 19.40 & 54.51\\
w/o Skill Optimization
    & 95.52 & 103.858 & 4.48 & 61.95\\
\bottomrule
\end{tabular*}

\end{table}

In this subsection, we assess the contributions of the key components of \method\ through ablation experiments on ALFWorld using DeepSeek-V4.1-Flash.
Specifically, we construct three variants by individually removing the Skill Skeleton, Programmer, and Skill Optimization.
As shown in Table~\ref{tab:ablation}, the variant without the Skill Skeleton achieves SR of 95.52\%, compared with 98.26\% for the full method.
Meanwhile, its token consumption increases by 2.75 times, and EIR rises from 2.99\% to 8.21\%, indicating the importance of explicit operation organization for efficient and reliable execution.
The most significant performance degradation occurs when the Programmer is disabled, with SR dropping to 53.48\%, token consumption increasing to 7.36 times that of the full method, and EIR reaching 19.40\%.
These results highlight the critical role of programmatic execution in improving task performance and reliability.
The variant without Skill Optimization achieves an SR of 95.52\%, while its token consumption increases to 2.48 times that of the full method.
Furthermore, APS decreases from 81.27\% to 61.95\%, demonstrating the importance of continuous skill Optimization for execution efficiency and consistency.
Overall, these results confirm the complementary contributions of the three components to \method's effectiveness.

\subsection{Discussion}
\para{Empirical Implications.} The performance gains vary across benchmarks, with larger improvements on LiveMath and OfficeQA than on SearchQA and DocVQA. 
This suggests that selective compilation is particularly beneficial when tasks contain recurring operations with explicit inputs, well-defined procedures, and verifiable effects. 
In contrast, tasks dominated by context-dependent interpretation offer fewer opportunities to replace LLM reasoning with executable procedures.
\para{Limitations.}\method\ relies on identifiable operation boundaries, observable preconditions, and independently checkable effects. 
Consequently, ambiguous or context-dependent decisions remain under LLM control, meaning that compilation reduces rather than eliminates execution uncertainty.
Moreover, local validation establishes correctness only under the tested inputs and environment states, without guaranteeing generalization to unseen conditions.
The benefits of persistent program reuse also depend on contract compatibility and sufficient reuse to amortize synthesis and validation costs. 
Finally, our evaluation of \method\ focused primarily on controlled benchmark environments, and its effectiveness across a broader range of tasks and in dynamic web environments remains to be investigated.

\balance
\section{Conclusion}\label{sec:conclusion}

\method\ integrates operation-level compilation into skill
evolution. Its executability-aware strategy combines validated
programs and context-dependent reasoning within hybrid
Skill Skeletons, while persistent program assets support
reuse and trace-driven Optimization.
Experiments across five benchmarks and six model families
show higher mean task performance. Fixed-bank repetitions
on LiveMath and ALFWorld additionally show fewer outcome
reversals. Resource savings vary by baseline, while ALFWorld
ablations identify the contributions of programmatic execution,
explicit operation dependencies, and Optimization.


\appendix
\section{Metric Definitions}\label{app:metrics}

\para{Effectiveness and reporting units.}
For method $m$, learning run $s$, and task $t$, let $z_{m,s,t}$ be the score returned by the benchmark evaluator, expressed on $[0,1]$ when the native score is reported as a percentage. The per-run score is
\begin{equation}
S_{m,s}=\frac{100}{|T|}\sum_{t\in T}z_{m,s,t}.
\end{equation}
EM and native binary task-success metrics have $z\in\{0,1\}$; any native continuous score keeps its own name and interpretation. The four-benchmark main table reports arithmetic means over independent learning runs. Its Avg column is the unweighted mean of SearchQA, OfficeQA, DocVQA, and LiveMath, and does not pool tasks across domains. ALFWorld success rate is reported separately. Repeated evaluations of one frozen bank form a separate replication axis. Secondary metrics are not converted into binary accuracy by an added threshold.

\para{Repeated outcomes.}
For a fixed bank and a binary correctness or success signal, let $y_{t,k}\in\{0,1\}$ be the outcome of repetition $k\in\{1,\ldots,K\}$ on task $t$. The following expressions are rates; percentages multiply them by 100:
\begin{align}
\RSR@K &=\frac{1}{|T|}\sum_{t\in T}\prod_{k=1}^{K}y_{t,k},\\
\EIR@K &=\frac{1}{|T|}\sum_{t\in T}\mathbf 1\!\left[0<\sum_{k=1}^{K}y_{t,k}<K\right].
\end{align}
Repeated-run EM or SR averages all $|T|K$ binary outcomes. EIR is the fraction of tasks exhibiting correctness flips; RSR is the supplementary fraction solved in every repetition. Both use tasks as their denominator, so the $K$ repetitions are not counted as independent tasks.

\para{Paths.}
Let $A_{t,k}$ be the canonical sequence of environment actions. Canonicalization removes method-internal bookkeeping, retains semantically relevant arguments, and expands environment calls inside programs. Restoration-discarded actions are excluded from the accepted path but remain in raw cost. With sequence Levenshtein distance $d$, define
\begin{align}
\operatorname{sim}(A,B)&=1-\frac{d(A,B)}{\max\{|A|,|B|,1\}},\\
\APS@K&=\frac{1}{|T|}\sum_{t\in T}\frac{2}{K(K-1)}\sum_{i<j}\operatorname{sim}(A_{t,i},A_{t,j}),\\
\EPR@K&=\frac{1}{|T|}\sum_{t\in T}\mathbf 1[A_{t,1}=\cdots=A_{t,K}].
\end{align}
These formulas are applied to workloads with a meaningful action sequence. A direct answer with no such sequence does not receive an artificial perfect path score. Success-conditioned APS averages pairs in which both executions succeed, first within each eligible task and then across eligible tasks.

\para{Resource variability.}
For nonnegative cost quantity $q_{t,k}$, let $\bar q_t$ and $s_t(q)$ be its mean and sample standard deviation across repetitions. For $\bar q_t>0$,
\begin{equation}
\CV_q(t)=100\,\frac{s_t(q)}{\bar q_t}.
\end{equation}
All-zero observed quantities have CV zero and are counted separately; missing quantities are excluded with their coverage stated. Mean task-level CV is reported with absolute mean usage. Failed episodes remain in the resource sample.

\para{Program execution and reuse.}
For seed $s$, let $T_s$ denote the test tasks and $K_s$ the frozen persistent Bank. Persistent Program Coverage (PPC) is the fraction of tasks in $T_s$ with at least one actually started invocation of a Program from $K_s$. A Program mentioned only in a plan, or synthesized temporarily for the current test task, does not contribute to PPC.

Consumed Program Outputs (CPO) reports a conservative, trace-confirmed lower bound on downstream consumption of Program results. A result contributes positive evidence only when its use by a subsequent operation, output binding, or accepted environment effect can be traced; a mere invocation or recorded return is insufficient. Cases without a verifiable consumption link are not inferred to be consumed. The reported lower bound is calculated within each seed before averaging.

Cross-Skill Reuse (CSR) is the fraction of actually used persistent Program versions that are invoked by at least two distinct Skill Skeletons. Multiple task instances of the same Skeleton do not by themselves establish cross-skill reuse. Program Invocation Success (PIS) is the number of invocations recorded as successful divided by all actually started Program invocations; invocation success alone does not establish final task correctness or universal semantic validity.

Program Action Share (PAS) is the ratio of canonical environment-action indices executed inside Program spans to all canonical environment-action indices. Program-internal environment actions are expanded, and the numerator is a union of indices so that nested bookkeeping does not double-count an action. Restoration-discarded actions are excluded from the accepted path but retained in cost accounting. Computation-only Programs with no environment action are captured by Program coverage rather than PAS.

For ALFWorld, the numerator/denominator pairs for CSR across seeds 42, 43, and 44 are $6/11$, $7/14$, and $6/10$; for PIS they are $690/807$, $641/713$, and $698/766$; and for PAS they are $1078/1209$, $979/1178$, and $1034/1163$, respectively. Each reported percentage is calculated per seed and then averaged equally across the three seeds. PPC is $134/134$ for each seed; the reported CPO values are lower bounds.

Graph Construction Coverage (GCC) is the fraction of tasks that enter an explicit nontrivial reusable or recomposed Skill Skeleton, excluding the one-node dynamic fallback. An LLM-free task requires no model request throughout the task execution; a Program identifier in the plan does not suffice.

\para{Statistical comparisons.}
Paired task-level differences use the same physical task identities. A task bootstrap resamples tasks while keeping their repeated executions together; it does not treat three executions of a task as three independent samples. Independent learned banks form a separate replication axis. Binary paired comparisons can use McNemar's test, with a stated multiplicity correction for a family of contrasts. No significance designation or uncertainty interval is inferred from rounded aggregate means alone.

\section{Validation and Execution Details}\label{app:validation}

\para{Source-grounded checks.}
A reusable program's test binding identifies the actual public inputs, source operation, and observed result. Required inputs are checked against the declared schema and linked to public fields or executed-operation arguments. Required outputs are linked to source result fields or independently recorded file effects. Hidden answer fields and task-identity lookup tables are not program inputs. An intermediate computation is not reclassified as a complete answer unless the recorded operation supports that output role.
For validation, a separate environment instance is reset to the source task and the relevant action prefix is replayed. The program runs through the permitted interface. The check compares required outputs with the source reference and, where applicable, verifies the native operation and its allowed writes. File checks compare the recorded output effect while retaining the original artifact hash. Parameter variations are applied only when the source supplies a valid independent reference and the varied input satisfies the schema. These tests are evidence of local behavior, not exhaustive branch coverage or semantic equivalence on arbitrary inputs.

\para{Side effects and failure handling.}
Isolated validation avoids changing the active task. Task-local execution uses a restorable snapshot where supported, and successful publication occurs only after the associated output checks. A failed attempt retains its actions, model usage, and error record. A restored-away interval is not positive evidence for later skill extraction. Where restoration is unavailable or fails, subsequent decisions depend on actual state observations; the framework does not assume arbitrary external operations are reversible.

\para{Retrieval and invocation qualification.}
In the ALFWorld execution protocol, Skill Skeleton and atomic-skill candidates are filtered for capability and contract compatibility before lexical ranking. Atomic skills use the Dice-based score in Eq.~\eqref{eq:retrieval_score}, including recorded utility and active status; no embedding-based retrieval is used. Program realizations are resolved through explicit Skill--Implementation--Program references rather than independently ranked by vector similarity. Eligibility requires input binding, preconditions, environment compatibility, and applicable validation evidence, as summarized in Eq.~\eqref{eq:program_compatibility}. When no unique eligible or preferred implementation is identified, execution remains agent-guided.

\para{Asset identity and compatibility.}
Operations, program realizations, and Skill Skeletons are
versioned separately.
Operations are consolidated only when they represent the
same local capability and have compatible contracts;
matching names or input--output types alone is insufficient.
Program reuse additionally requires compatibility with the
current inputs, preconditions, expected effects, and execution
environment.
The Program Asset Manager associates each accepted program
version with its operation contract, applicable execution
conditions, and validation evidence.
Changes to the implementation, behavioral contract, or
supported execution conditions require validation of the
affected realization before admission.
A failed realization does not invalidate the operation
definition itself.
Compatible replacements can serve existing Skill Skeletons if their operation requirements and parameter bindings remain satisfied.

\section{Learning, Selection, and Reproducibility}
\label{app:protocol}

\para{Training and trace-based learning.}
Training trajectories provide evidence for skill induction
and revision.
The batchwise evolution process described in
Section~\ref{sec:method} allows accepted updates to influence
subsequent task execution.
The Skill Optimizer analyzes both successful and failed
episodes to propose local changes and consolidate compatible
skills.
Retained training trajectories remain associated with the
skill state and environmental observations of their original
execution when reused for further revision.
Task outcomes and local operation outcomes are distinguished.
An unsuccessful task may contain independently checked
successful operations that support skill learning.
Conversely, overall task success does not establish that
every intermediate program executed correctly.
Failed calls and discarded or rolled-back segments support
diagnosis and revision, while only independently checked,
retained execution segments provide positive evidence for
reusable operations.

\para{Local validation and candidate selection.}
Local program validation checks whether a candidate realizes
its operation under the tested inputs and execution conditions,
using reference evidence independent of the candidate's
claimed outputs.
Validation-set performance is used to select among candidate
skills or bank snapshots.
This selection evaluates their contribution to task
performance and does not replace local program validation.
The selected persistent skill state is fixed before the
corresponding test evaluation.

\para{Independent learning runs.}
Main task results and cost comparisons use seeds
$42$, $43$, and $44$.
Each learning-based method constructs a separate bank for
each model--benchmark--seed configuration.
Task partitions are shared across methods and runs.
Main performance and cost metrics are calculated within each
independent run and then averaged over the three seeds.
Fixed-bank repetition metrics are reported separately.
Learning seeds control training order and supported local
sampling; they do not imply deterministic responses from
a hosted language model.
No Skill and Human Skill are evaluated under the same
run settings without model-based skill learning.

\para{Repeated execution.}
Stability and ablation analyses use the skill state learned
with seed $42$.
For each method or condition, the same test tasks are executed
three times with fixed persistent skills, task inputs,
initial environment states, and inference settings.
These repetitions examine variation in executing an
established skill rather than variation between independently
learned banks.
The LiveMath analysis uses Gemini-3.8-Flash on 100 tasks,
giving 300 task executions.
The ALFWorld analysis uses DeepSeek-V4.1-Flash on 134 tasks,
giving 402 task executions.
Mean EM or SR summarizes execution outcomes, whereas EIR@3
counts tasks with both successful and unsuccessful outcomes.
ALFWorld path metrics use canonical environment-action
sequences, including actions executed inside programs.
The Full ablation condition uses the same skill bank and
repeated evaluations as the complete framework in the
ALFWorld stability analysis.
Metric definitions are given in Appendix~\ref{app:metrics}.

\para{Frozen skills and task-local execution.}
The persistent learned skill and Program Asset Bank state remains
fixed throughout test evaluation.
Task-local argument grounding, error recovery, and programming
permitted by the corresponding method remain confined to the
current task execution.
Temporary programs and intermediate results are not added
to the persistent bank or transferred to later test tasks.
Each repeated execution starts from the same selected bank
and reset task-local state.
Test outcomes are used for reporting performance and
execution behavior, not for selecting or revising the bank.

\para{Baselines and textual priors.}
Human Skill uses the benchmark-specific human-written
\texttt{initial.md} released by the SkillOpt
authors~\citep{skillopt}.
This textual skill remains unchanged during testing and
repeated execution.
Other baselines retain their own initialization procedures,
learning algorithms, and knowledge representations under
the shared task interface.
Within each model--benchmark comparison, methods use the
same task resources, native tools, evaluators, and inference
settings.
The reported \method\ banks are learned through the skill
evolution process described in Section~\ref{sec:method}.

\para{Ablation scope.}
The three interventions in Section~\ref{sec:ablation} are
evaluated on ALFWorld with DeepSeek-V4.1-Flash.
Each condition learns independently from an empty bank
using seed $42$, and its intervention applies during both
learning and testing.

Without Skill Skeleton disables multi-operation skeleton
construction and retrieval while retaining individual
operation retrieval, programmatic execution, and trace-driven
learning.
Without Programmer disables program synthesis, qualification,
and invocation, including reuse of stored program realizations,
while retaining textual skill learning, Skill Skeletons,
and native tools.
Without Skill Optimization permits the acquisition of new
capabilities and an existing textual operation's first
validated program realization, but prevents subsequent
semantic revision or merging of accepted guidance, contracts,
skeletons, and program realizations.
Initial program validation and bounded repair remain available.
All conditions retain runtime grounding and error recovery
within their permitted execution mechanisms, with shared
task partitions, native tools, and budget limits.

\para{Resource accounting.}
Learning tokens include training-trajectory collection,
skill construction and revision, feedback, and validation-based
selection.
Each model request is counted once, and each bank's learning
investment is counted once before averaging across learning
runs.
Test tokens include input and output tokens from successful
and unsuccessful executions, method-internal model calls,
and retries.
Calls/task counts actual model HTTP requests.
Retries contribute to the cost of their original task
execution without increasing the number of evaluated tasks.
An execution completed entirely by programs contributes
zero model requests.
The main cost tables average the per-run measurements from
seeds $42$, $43$, and $44$.
Ablation costs instead average the three repeated evaluations
of each condition's seed-42 bank.
These summaries characterize resource use across independently
learned banks and during repeated use of a fixed bank,
respectively.

\para{Reproducibility and development history.}
Experimental records associate each run with its model and
inference settings, task partition, learning seed, selected
bank version, task outcomes, execution traces, and learning
and test resource usage.
Reasoning settings, completion limits, and retry policies
are held consistent within each model--benchmark comparison.
Development included inspection of held-out failure cases;
the reported frozen-bank evaluations prohibit persistent
updates from test outcomes.

\para{Scope of validation.}
Programs execute within the permitted benchmark interfaces
and task resources under bounded execution.
Local validation provides evidence for the checked inputs,
environment configurations, and operation effects.
Its coverage is limited to those conditions.
Task correctness is determined by the benchmark evaluator,
while unresolved semantics and context-dependent decisions
remain under the target agent's control.




\begin{thebibliography}{37}

\bibitem{react}
Shunyu Yao, Jeffrey Zhao, Dian Yu, Nan Du, Izhak Shafran, Karthik Narasimhan, and Yuan Cao. 2023.
ReAct: Synergizing Reasoning and Acting in Language Models.
In \emph{International Conference on Learning Representations}.
\url{https://arxiv.org/abs/2210.03629}

\bibitem{mind2web}
Xiang Deng, Yu Gu, Boyuan Zheng, Shijie Chen, Samuel Stevens, Boshi Wang, Huan Sun, and Yu Su. 2023.
Mind2Web: Towards a Generalist Agent for the Web.
In \emph{Advances in Neural Information Processing Systems},
Vol. 36.
\url{https://arxiv.org/abs/2306.06070}

\bibitem{webarena}
Shuyan Zhou, Frank F. Xu, Hao Zhu, Xuhui Zhou, Robert Lo, Abishek Sridhar, Xianyi Cheng, Tianyue Ou, Yonatan Bisk, Daniel Fried, Uri Alon, and Graham Neubig. 2024.
WebArena: A Realistic Web Environment for Building Autonomous Agents.
In \emph{International Conference on Learning Representations}.
\url{https://arxiv.org/abs/2307.13854}

\bibitem{webvoyager}
Hongliang He, Wenlin Yao, Kaixin Ma, Wenhao Yu, Yong Dai, Hongming Zhang, Zhenzhong Lan, and Dong Yu. 2024.
WebVoyager: Building an End-to-End Web Agent with Large Multimodal Models.
In \emph{Proceedings of the 62nd Annual Meeting of the Association for Computational Linguistics}, 6864--6890.
\url{https://doi.org/10.18653/v1/2024.acl-long.371}

\bibitem{autotools}
Zhengliang Shi, Shen Gao, Lingyong Yan, Yue Feng, Xiuyi Chen, Zhumin Chen, Dawei Yin, Suzan Verberne, and Zhaochun Ren. 2025.
Tool Learning in the Wild: Empowering Language Models as Automatic Tool Agents.
In \emph{Proceedings of the ACM Web Conference 2025}.
\url{https://doi.org/10.1145/3696410.3714825}

\bibitem{reflexion}
Noah Shinn, Federico Cassano, Ashwin Gopinath, Karthik Narasimhan, and Shunyu Yao. 2023.
Reflexion: Language Agents with Verbal Reinforcement Learning.
In \emph{Advances in Neural Information Processing Systems}, Vol. 36.
\url{https://papers.neurips.cc/paper_files/paper/2023/hash/1b44b878bb782e6954cd888628510e90-Abstract-Conference.html}

\bibitem{expel}
Andrew Zhao, Daniel Huang, Quentin Xu, Matthieu Lin, Yong-Jin Liu, and Gao Huang. 2024.
ExpeL: LLM Agents Are Experiential Learners.
\emph{Proceedings of the AAAI Conference on Artificial Intelligence} 38, 17 (2024), 19632--19642.
\url{https://doi.org/10.1609/aaai.v38i17.29936}


\bibitem{fedskillgit}
Boxun Lan, Jiamei Li, Weijian Li, Chengyu Yang, Zhuang Qi, Chengcheng Wan, Xiaohong Chen, and Ming Hu. 2026.
FedSkillGit: Federated Skill Evolution via Version Control.
\emph{arXiv:2610.21348}.
\url{https://arxiv.org/abs/2610.21348}

\bibitem{trace2skill}
Jingwei Ni, Yihao Liu, Xinpeng Liu, Yutao Sun, Mengyu Zhou, Pengyu Cheng, Dexin Wang, Erchao Zhao, Xiaoxi Jiang, and Guanjun Jiang. 2026.
Trace2Skill: Distill Trajectory-Local Lessons into Transferable Agent Skills.
\emph{arXiv:2603.25158}.
\url{https://arxiv.org/abs/2603.25158}

\bibitem{skillopt}
Yifan Yang, Ziyang Gong, Weiquan Huang, Qihao Yang, Ziwei Zhou, Zisu Huang, Yan Li, Xuemei Gao, Qi Dai, Bei Liu, Kai Qiu, Yuqing Yang, Dongdong Chen, Xue Yang, and Chong Luo. 2026.
SkillOpt: Executive Strategy for Self-Evolving Agent Skills.
\emph{arXiv:2605.23904}.
\url{https://arxiv.org/abs/2605.23904}

\bibitem{gepa}
Lakshya A. Agrawal, Shangyin Tan, Dilara Soylu, Noah Ziems, Rishi Khare, Krista Opsahl-Ong, Arnav Singhvi, Herumb Shandilya, Michael J. Ryan, Meng Jiang, Christopher Potts, Koushik Sen, Alex Dimakis, Ion Stoica, Dan Klein, Matei Zaharia, and Omar Khattab. 2026.
GEPA: Reflective Prompt Evolution Can Outperform Reinforcement Learning.
In \emph{International Conference on Learning Representations}.
\url{https://proceedings.iclr.cc/paper_files/paper/2026/hash/0e9e708b6f48e14fd0ac29e167413f76-Abstract-Conference.html}

\bibitem{embodiskill}
Ruofei Ju, Xinrui Wang, Xin Ding, Yifan Yang, Hao Wu, Shiqi Jiang, Qianxi Zhang, Hao Wen, Xiangyu Li, Weijun Wang, Kun Li, Yunxin Liu, Haipeng Dai, Wei Wang, and Ting Cao. 2026.
EmbodiSkill: Skill-Aware Reflection for Self-Evolving Embodied Agents.
\emph{arXiv:2605.10332}.
\url{https://arxiv.org/abs/2605.10332}

\bibitem{evoskill}
Salaheddin Alzubi, Noah Provenzano, Jaydon Bingham, Weiyuan Chen, and Tu Vu. 2026.
EvoSkill: Automated Skill Discovery for Multi-Agent Systems.
\emph{arXiv:2603.02766}, version 1.
\url{https://arxiv.org/abs/2603.02766v1}

\bibitem{voyager}
Guanzhi Wang, Yuqi Xie, Yunfan Jiang, Ajay Mandlekar, Chaowei Xiao, Yuke Zhu, Linxi Fan, and Anima Anandkumar. 2024.
Voyager: An Open-Ended Embodied Agent with Large Language Models.
\emph{Transactions on Machine Learning Research}.
\url{https://openreview.net/forum?id=Mn3Lzp1sMQ}

\bibitem{polyskill}
Simon Yu, Gang Li, Weiyan Shi, and Peng Qi. 2026. 
PolySkill: Learning Generalizable Skills Through Polymorphic Abstraction for Continual Learning.
In \emph{International Conference on Learning Representations}.
\url{https://proceedings.iclr.cc/paper_files/paper/2026/hash/e350897ed832f9cad6f2e223e7acad6d-Abstract-Conference.html}

\bibitem{flowevo}
Zeyu Ren, Ling Yue, Ran Li, Yishu Wang, Shengxiang Xu, Hanmo Liu, Shaowu Pan, and Shimin Di. 2026.
FlowEvo: Self-Evolving Agents through the Co-Evolution of Workflows and Executable Skills.
In \emph{Third Conference on Language Modeling}.
\url{https://arxiv.org/abs/2607.21596}

\bibitem{awm}
Zora Zhiruo Wang, Jiayuan Mao, Daniel Fried, and Graham Neubig. 2025.
Agent Workflow Memory.
In \emph{Proceedings of the 42nd International Conference on Machine Learning}, PMLR 267, 63897--63911.
\url{https://proceedings.mlr.press/v267/wang25bx.html}

\bibitem{skillweaver}
Boyuan Zheng, Michael Y. Fatemi, Xiaolong Jin, Zora Zhiruo Wang, Apurva Gandhi, Yueqi Song, Yu Gu, Jayanth Srinivasa, Gaowen Liu, Graham Neubig, and Yu Su. 2025.
SkillWeaver: Web Agents Can Self-Improve by Discovering and Honing Skills.
\emph{arXiv:2504.07079}.
\url{https://arxiv.org/abs/2504.07079}

\bibitem{asi}
Zora Zhiruo Wang, Apurva Gandhi, Graham Neubig, and Daniel Fried. 2025.
Inducing Programmatic Skills for Agentic Tasks.
In \emph{Second Conference on Language Modeling}.
\url{https://openreview.net/forum?id=lsAY6fWsog}

\bibitem{nsi}
Jie-Jing Shao, Haiyan Yin, Yueming Lyu, Xingrui Yu, Lan-Zhe Guo, Ivor Tsang, James Kwok, and Yu-Feng Li. 2026.
Lifting Traces to Logic: Programmatic Skill Induction with Neuro-Symbolic Learning for Long-Horizon Agentic Tasks.
In \emph{Proceedings of the 43rd International Conference on Machine Learning}, PMLR 306, 110314--110337.
\url{https://proceedings.mlr.press/v306/shao26i.html}

\bibitem{skilldisco}
Zhongxin Guo, Danrui Qi, Hanwen Gu, Peng Cheng, and Yongqiang Xiong. 2026.
SKILL-DISCO: Distilling and Compiling Agent Traces into Reusable Procedural Skills.
\emph{arXiv:2606.26669}.
\url{https://arxiv.org/abs/2606.26669}

\bibitem{pal}
Luyu Gao, Aman Madaan, Shuyan Zhou, Uri Alon, Pengfei Liu, Yiming Yang, Jamie Callan, and Graham Neubig. 2023.
PAL: Program-aided Language Models.
In \emph{Proceedings of the 40th International Conference on Machine Learning}.
\url{https://arxiv.org/abs/2211.10435}

\bibitem{pot}
Wenhu Chen, Xueguang Ma, Xinyi Wang, and William W. Cohen. 2023.
Program of Thoughts Prompting: Disentangling Computation from Reasoning for Numerical Reasoning Tasks.
\emph{Transactions on Machine Learning Research}.
\url{https://arxiv.org/abs/2211.12588}

\bibitem{llmcompiler}
Sehoon Kim, Suhong Moon, Ryan Tabrizi, Nicholas Lee, Michael W. Mahoney, Kurt Keutzer, and Amir Gholami. 2024.
An LLM Compiler for Parallel Function Calling.
In \emph{Proceedings of the 41st International Conference on Machine Learning}, PMLR 235, 24370--24391.
\url{https://proceedings.mlr.press/v235/kim24y.html}

\bibitem{dspy}
Omar Khattab, Arnav Singhvi, Paridhi Maheshwari, Zhiyuan Zhang, Keshav Santhanam, Sri Vardhamanan A, Saiful Haq, Ashutosh Sharma, Thomas Joshi, Hanna Moazam, Heather Miller, Matei Zaharia, and Christopher Potts. 2024.
DSPy: Compiling Declarative Language Model Calls into State-of-the-Art Pipelines.
In \emph{International Conference on Learning Representations}.
\url{https://proceedings.iclr.cc/paper_files/paper/2024/hash/f1cf02ce09757f57c3b93c0db83181e0-Abstract-Conference.html}

\bibitem{skillx}
Chenxi Wang, Zhuoyun Yu, Xin Xie, Wuguannan Yao, Runnan Fang, Shuofei Qiao, Kexin Cao, Guozhou Zheng, Xiang Qi, Peng Zhang, and Shumin Deng. 2026.
SkillX: Automatically Constructing Skill Knowledge Bases for Agents.
\emph{arXiv:2604.04804}.
\url{https://arxiv.org/abs/2604.04804}

\bibitem{codepolicies}
Jacky Liang, Wenlong Huang, Fei Xia, Peng Xu, Karol Hausman, Brian Ichter, Pete Florence, and Andy Zeng. 2023.
Code as Policies: Language Model Programs for Embodied Control.
In \emph{IEEE International Conference on Robotics and Automation}.
\url{https://arxiv.org/abs/2209.07753}

\bibitem{searchqa}
Matthew Dunn, Levent Sagun, Mike Higgins, V. Ugur Guney, Volkan Cirik, and Kyunghyun Cho. 2017.
SearchQA: A New Q\&A Dataset Augmented with Context from a Search Engine.
\emph{arXiv:1704.05179}.
\url{https://arxiv.org/abs/1704.05179}

\bibitem{officeqa}
Databricks AI Research Team. 2025. 
Introducing OfficeQA: A Benchmark for End-to-End Grounded Reasoning.
\emph{Databricks Research}, December 9, 2025.
\url{https://www.databricks.com/blog/introducing-officeqa-benchmark-end-to-end-grounded-reasoning}

\bibitem{docvqa}
Minesh Mathew, Dimosthenis Karatzas, and C. V. Jawahar. 2021.
DocVQA: A Dataset for VQA on Document Images. 
In \emph{IEEE Winter Conference on Applications of Computer Vision}, 2200--2209.
\url{https://arxiv.org/abs/2007.00398}

\bibitem{livemath}
Linyang He, Qiyao Yu, Hanze Dong, Baohao Liao, Xinxing Xu, Micah Goldblum, Jiang Bian, and Nima Mesgarani. 2026.
LiveMathematicianBench: A Live Benchmark for Mathematician-Level Reasoning with Proof Sketches.
\emph{arXiv:2604.01754}, version 1.
\url{https://arxiv.org/abs/2604.01754v1}

\bibitem{alfworld}
Mohit Shridhar, Xingdi Yuan, Marc-Alexandre C\^ot\'e, Yonatan Bisk, Adam Trischler, and Matthew Hausknecht. 2021.
ALFWorld: Aligning Text and Embodied Environments for Interactive Learning.
In \emph{International Conference on Learning Representations}.
\url{https://arxiv.org/abs/2010.03768}

\bibitem{toolformer}
Timo Schick, Jane Dwivedi-Yu, Roberto Dessi, Roberta Raileanu, Maria Lomeli, Eric Hambro, Luke Zettlemoyer, Nicola Cancedda, and Thomas Scialom. 2023.
Toolformer: Language Models Can Teach Themselves to Use Tools.
In \emph{Advances in Neural Information Processing Systems}, Vol. 36.
\url{https://papers.nips.cc/paper_files/paper/2023/hash/d842425e4bf79ba039352da0f658a906-Abstract-Conference.html}

\bibitem{creator}
Cheng Qian, Chi Han, Yi Fung, Yujia Qin, Zhiyuan Liu, and Heng Ji. 2023.
CREATOR: Tool Creation for Disentangling Abstract and Concrete Reasoning of Large Language Models.
In \emph{Findings of the Association for Computational Linguistics: EMNLP 2023}, 6922--6939.
\url{https://doi.org/10.18653/v1/2023.findings-emnlp.462}

\bibitem{latm}
Tianle Cai, Xuezhi Wang, Tengyu Ma, Xinyun Chen, and Denny Zhou. 2024.
Large Language Models as Tool Makers.
In \emph{International Conference on Learning Representations}.
\url{https://arxiv.org/abs/2305.17126}

\bibitem{lilo}
Gabriel Grand, Lio Wong, Maddy Bowers, Theo X. Olausson, Muxin Liu, Joshua B. Tenenbaum, and Jacob Andreas. 2024.
LILO: Learning Interpretable Libraries by Compressing and Documenting Code.
In \emph{International Conference on Learning Representations}.
\url{https://proceedings.iclr.cc/paper_files/paper/2024/hash/819cebb05f993840e8a52d7564c5c282-Abstract-Conference.html}

\bibitem{legoprover}
Haiming Wang, Huajian Xin, Chuanyang Zheng, Zhengying Liu, Qingxing Cao, Yinya Huang, Jing Xiong, Han Shi, Enze Xie, Jian Yin, Zhenguo Li, and Xiaodan Liang. 2024.
LEGO-Prover: Neural Theorem Proving with Growing Libraries.
In \emph{International Conference on Learning Representations}.
\url{https://proceedings.iclr.cc/paper_files/paper/2024/hash/85dca46374dc0f27b4bb5f265b3d17f0-Abstract-Conference.html}

\bibitem{dsagent}
Siyuan Guo, Cheng Deng, Ying Wen, Hechang Chen, Yi Chang, and Jun Wang. 2024.
DS-Agent: Automated Data Science by Empowering Large Language Models with Case-Based Reasoning.
In \emph{Proceedings of the 41st International Conference on Machine Learning}, PMLR 235, 16813--16848.
\url{https://proceedings.mlr.press/v235/guo24b.html}

\end{thebibliography}
\end{document}